\documentclass[9pt]{extarticle}
\usepackage{spconf,amsmath,graphicx}
\usepackage{cite}
\usepackage{bm}
\usepackage{amssymb,amsfonts}
\usepackage{algorithm}
\usepackage{algorithmic}
\usepackage{textcomp}
\usepackage{xcolor}

\usepackage{enumitem}
\setlist{nosep, leftmargin=14pt}

\newcommand{\figref}[1]{Fig.~\ref{#1}}

\newcommand{\tabref}[1]{Table~\ref{#1}}
\newcommand{\algoref}[1]{Algorithm~\ref{#1}}

\title{\fontsize{14pt}{14pt}\selectfont SAM-Correction: Fully Adaptive Label Noise Reduction for Medical Image Segmentation}
\name{Takuro Shimaya \qquad Masahiro Saiko}
\address{Biometrics Research Laboratories, NEC Corporation, Kawasaki, Japan}
\begin{document}
%
\maketitle
\begingroup
\renewcommand{\thefootnote}{}
\footnotetext{%
\textcopyright~2024 IEEE. Personal use of this material is permitted.
Permission from IEEE must be obtained for all other uses,
in any current or future media, including reprinting/republishing
this material for advertising or promotional purposes,
creating new collective works, for resale or redistribution
to servers or lists, or reuse of any copyrighted component
of this work in other works.
}
\endgroup
\begin{abstract}
The quality of the teacher label has a dramatic impact on the outcome of the machine learning process.
Medical image segmentation often suffers from a noisy ground truth (GT) mask due to difficulties in consistent annotation.
Although a recent study has proposed a promising method to curate the noisy GT mask, this requires additional well-curated GT masks that are not always accessible without the aid of multiple medical experts.
As an alternative, we propose SAM-Correction, which utilizes a foundation model to correct the noisy GT mask in a fully adaptive manner instead of manually preparing well-curated masks.
In our approach, we progressively mitigate label noise via the segment anything model (SAM), where prompts are given from another segmentation network that we train concurrently.
Experiments with various types of label noise derived from actual human annotations demonstrate that SAM-C achieves precise noise reduction and high-performing segmentation models trained thereby.
Our method paves the way for robust model training even when clean labels are entirely unavailable.
\end{abstract}
\section{Introduction}
Noisy mask annotations are known to degrade the performance of deep neural networks (DNN) for image segmentation tasks.
This problem is particularly evident in medical images, where noise can originate from a lack of uniqueness in object boundaries, the necessity of expertise, or the limited time that annotators can devote to the task \cite{Zhang2020,Yao2023}.
Label noise problems themselves have historically been tackled in the context of classification tasks \cite{Wang2019,Liu2020}.
While this approach is applicable to image segmentation by assuming an independent classification problem for each pixel, its contribution is limited due to lacking consideration of the spatial correlation of noise \cite{Yao2023}.
At present, strategies dedicated to the image segmentation context have made relatively little progress.

Yao \textit{et al.} recently proposed a method specific to noisy mask annotations based on a Markov model \cite{Yao2023}.
In their method, spatial correction (SC) is utilized to correct the noisy ground truth (GT) mask for training a DNN, assuming that noisy GT masks are biased, i.e., that they expand or shrink from clean GT masks by a typical length.
The typical length is defined as a bias parameter and estimated by comparing the prediction from a DNN trained by noisy GTs with the corresponding well-curated clean GT prepared manually.
In this paper, we refer to the above process as true validation (TV). 
With TV using additionally prepared clean GTs, SC has outperformed existing methods while leaving the following concern.
For medical images, clean GTs are typically determined by majority voting among several noisy GTs annotated by each expert \cite{Zhang2020}. 
Such clean GTs reflecting consensus among multiple experts are not always accessible due to the high annotation cost.
Moreover, SC would be sub-optimal because it attempts to treat noise uniformly with the single bias parameter.

Instead of preparing additional clean GTs, we utilize a foundation model to recover clean labels only from noisy GTs.
Specifically, we adopt the segment anything model (SAM) \cite{kirillov2023}, which achieves zero-shot segmentation by the user's prompts (e.g., boxes and masks) to specify which object to segment in a given image.
With the aid of SAM, our proposed method, SAM-Correction (SAM-C), iteratively reduces label noise along with training a segmentation DNN.
SAM-C aims to reproduce clean GTs akin to those obtained by majority voting, where SAM generates the most reasonable mask using generic image features and prompts from the trained DNN. 
Unlike SC, our method allows fully adaptive noise correction without additional clean GTs and the bias parameter.
Our experiments using two datasets with non-synthetic noise yielded the following insights:
\begin{enumerate}
\item SAM-C outperformed SC in terms of the accuracy of segmentation models under noisy teacher labels.
\item SAM-C achieved precise noise reduction from low-quality annotations, while SC failed in some cases.
\item Our fully adaptive framework does not require additional well-curated GTs, and is applicable even when one cannot enlist the help of multiple clinical experts.
\end{enumerate}

\section{Methods}
\begin{figure*}[t]
    \centerline{\includegraphics[width=\linewidth]{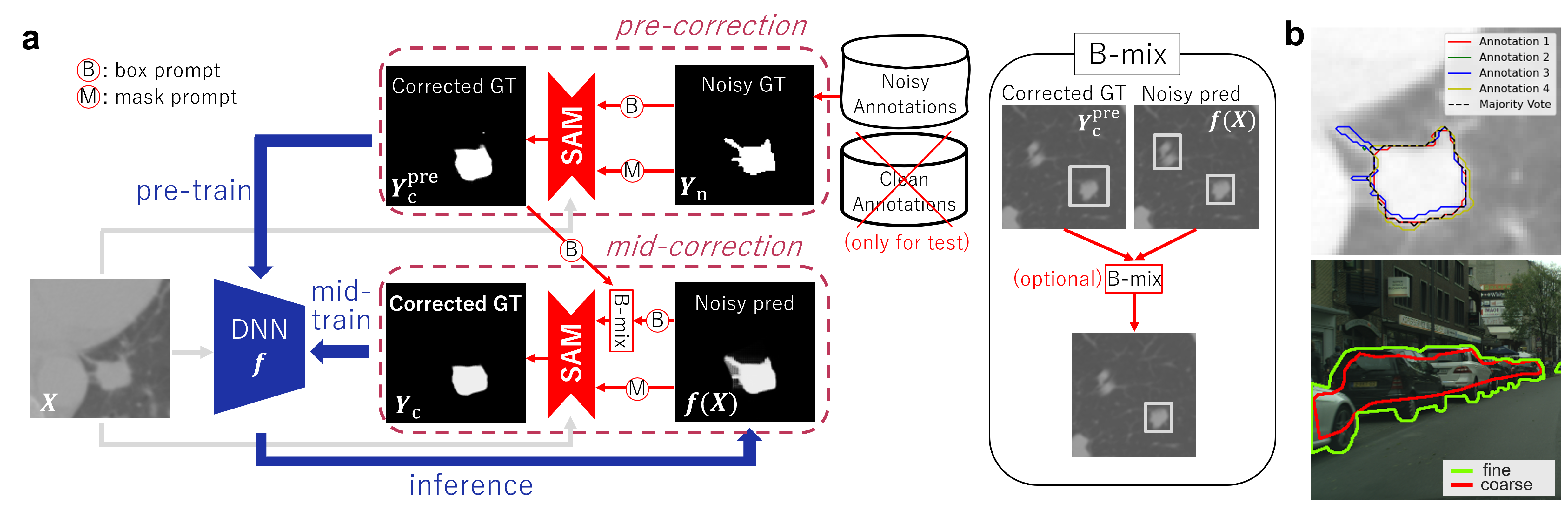}}
    \caption{Proposed framework and data utilized in this study.
    (a) Overview of SAM-C (see the main text for details).
    B-mix excludes the bounding box from a noisy prediction that does not overlap with any of those in $Y_\mathrm{c}^\mathrm{pre}$.
    The B-mix gate can be disabled to directly use prompts from the noisy prediction.
    The post-correction is not shown here.
    (b) Examples from LIDC-IDRI (top) and Cityscapes (bottom).
    }
    \label{fig:1}
\end{figure*}

\subsection{SAM-C}
The overview of SAM-C is shown in \figref{fig:1}(a) and \algoref{algo:1}.
We assume that one can only access the dataset $\mathcal{D}_\mathrm{n}$ composed of noisy GT masks $\bm{Y}_\mathrm{n}$ and corresponding images $\bm{X}$.
The aim of SAM-C is to obtain corrected GT masks $\bm{Y}_\mathrm{c}$ and train a high-performing segmentation DNN $\bm{f}$ (U-Net \cite{Ronneberger2015} in this study).
Our approach consists of two key processes, pre-correction and mid-correction explained below.
The specific outcomes from each process are discussed in Sec.~III.

The first step is to correct noisy GT masks $\bm{Y}_\mathrm{n}$ with the aid of SAM, which we call ``pre-correction''.
We extracted bounding boxes for each object, i.e., each connected component in $\bm{Y}_\mathrm{n}$.
These boxes and the mask were then utilized as prompts for SAM to obtain the corrected GT mask $\bm{Y}_\mathrm{c}^\mathrm{pre}$.
We then trained a DNN $\bm{f}$ using $\bm{Y}_\mathrm{c}^\mathrm{pre}$ (``pre-train'' in \figref{fig:1}(a)).
\begin{figure}[t]
    \begin{algorithm}[H]
        \caption{SAM-C (with pre- and mid-correction)}
        \label{algo:1}
        \begin{algorithmic}[1]
        \REQUIRE Training dataset $\mathcal{D}_\mathrm{n}$ with images $\bm{X}$ and noisy GT masks $\bm{Y}_\mathrm{n}$, no. of cycles $N_\mathrm{c}$, and whether to use B-mix.
        \ENSURE Corrected GT masks $\bm{Y}_\mathrm{c}$ and an accurate DNN $\bm{f}$.
        \STATE Curate noisy GT masks $\bm{Y}_\mathrm{n}$ by SAM and get corrected GT masks $\bm{Y}_\mathrm{c}^\mathrm{pre}$ (pre-correction).
        \STATE Train $\bm{f}$ using $\bm{Y}_\mathrm{c}^\mathrm{pre}$ obtained by the pre-correction.
        \FOR{$i=1$ to $N_\mathrm{c}$}
        \STATE Get noisy prediction masks $\bm{f}(\bm{X})$.
        \IF{B-mix is enabled}
        \STATE Get prompts via B-mix from $\bm{Y}_\mathrm{c}^\mathrm{pre}$ and $\bm{f}(\bm{X})$.
        \ELSE
        \STATE Get prompts directly from $\bm{f}(\bm{X})$.
        \ENDIF
        \STATE Get the corrected GT masks $\bm{Y}_\mathrm{c}$ (mid-correction).
        \STATE Train $\bm{f}$ using $\bm{Y}_\mathrm{c}$ obtained by the mid-correction.
        \ENDFOR
        \RETURN $\bm{Y}_\mathrm{c}$, $\bm{f}$
        \end{algorithmic}
    \end{algorithm}
\end{figure}

The next step is to further curate the GT mask through ``mid-correction'' as follows.
We first get bounding boxes and the mask from the noisy prediction $\bm{f}(\bm{X})$, the same as in the pre-correction.
Although we can directly use these prompts to generate the corrected GT, this may cause hallucinations due to false positives by the DNN.
Here, we incorporate a B-mix gate to avoid such hallucination, as depicted in \figref{fig:1}(a).
The B-mix gate excludes bounding boxes from $\bm{f}(\bm{X})$ that do not overlap with those from $Y_\mathrm{c}^\mathrm{pre}$.
We then obtain the corrected GT mask $\bm{Y}_\mathrm{c}$ by using the remaining bounding boxes and the noisy prediction $\bm{f}(\bm{X})$ as prompts for SAM.
Note that the B-mix gate can be disabled to directly get prompts from $\bm{f}(\bm{X})$, especially when hallucination is desirable for mask correction.
The DNN $\bm{f}$ is then re-trained using $\bm{Y}_\mathrm{c}$ (mid-train in \figref{fig:1}(a)).
As shown in \algoref{algo:1}, the mid-correction can be repeated any $N_\mathrm{c}$ times.
Consequently, SAM-C has two unique hyperparameters to be determined by validation: the number of cycles $N_\mathrm{c}$ and whether to use the B-mix gate.
After $N_\mathrm{c}$ cycles of the mid-correction, we finally get well-corrected GTs $\bm{Y}_\mathrm{c}$ and a DNN $\bm{f}$ predicting clean masks.

We should remark that the mid-correction is partly inspired by SC, in which the corrected GT $\bm{Y}_\mathrm{c}$ is analytically calculated from the noisy prediction $\bm{f}(\bm{X})$ according to the bias parameter.
The potential limitation of SC is the estimation of the bias parameter via TV, which requires manually well-curated GT masks.
In contrast, SAM-C can derivate $\bm{Y}_\mathrm{c}$ by utilizing SAM instead of TV.

\subsection{Datasets}
Following Yao \textit{et al.} \cite{Yao2023}, we utilized the LIDC-IDRI \cite{Armato2011} and Cityscapes \cite{Cordts2016} datasets, both of which contain non-synthetic masks, i.e., human annotations with actual noise.
We avoided using datasets modified with artificial noise utilized in ref.\cite{Yao2023} to focus on efficacy against actual noise.
LIDC-IDRI contains thorax CT scans annotated by four radiologists (see \figref{fig:1}(b), top).
Following prior studies \cite{Yao2023,Kohl2018}, we extracted 2D square slices centered around nodules.
The clean GT mask for test and TV was obtained by majority voting.
The noisy mask for training and validation was defined as ``or'', ``random select'', or ``worst''.
We obtained the ``or'' noisy GT mask as the union of four annotations (as done by Yao \textit{et al.} \cite{Yao2023}), which becomes more widespread than the clean GT.
The ``random select'' mask was randomly selected from four annotations, corresponding to when data are divided among several annotators, but only one person annotates each datum.
The ``worst'' mask was defined as one of four annotations with the lowest Dice similarity coefficient (DSC) with the clean GT mask.
We finally obtained 9,335 images for train, 1,760 for validation, and 1,990 for test.

The Cityscapes dataset contains 5,000 images with fine and coarse annotations (see \figref{fig:1}(b), bottom).
The coarse annotation lacks the detailed structure near the margin, which is also present in medical image datasets (see \figref{fig:1}(b) and \figref{fig:2} for annotations in the LIDC-IDRI).
We again followed Yao \textit{et al.} \cite{Yao2023} to preprocess the data, resulting in 2,405 images for train, 130 for validation, and 479 for test.
\begin{table*}[t]
  \caption{DSC between the prediction of $\bm{f}$ and the clean GT.
  The average and the standard error over five independent trials are shown.
  (a) Comparison of our method with others.
  (b) Ablation study for pre-, mid-, and post-correction processes.}
  \label{tab:1}
  \centerline{\includegraphics[width=0.75\linewidth]{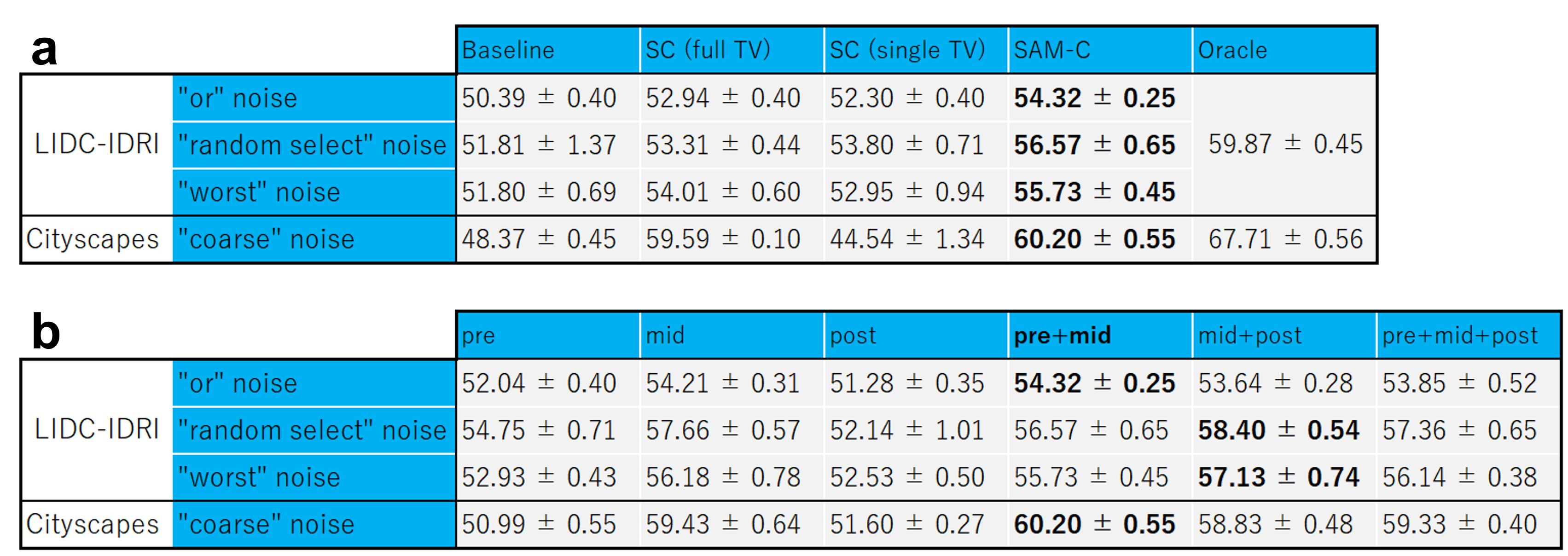}}
\end{table*}

\subsection{Experiments}
We compared the performance of SAM-C with the baseline, SC with full TV, SC with single TV, and the oracle.
Regarding the baseline, we trained the U-Net $\bm{f}$ only with the noisy GT mask $\bm{Y}_\mathrm{n}$.
SC with full TV used all clean GTs in the validation dataset to estimate the bias parameter, while SC with single TV used a single randomly selected clean GT.
The performance of SC was re-evaluated with hyperparameter tuning by ourselves.
For the oracle, instead of noisy GTs, clean GTs were fully used for training.
Note that the oracle is not for simple comparison but rather for estimating the upper bound of improvement.
The above five methods were examined under four conditions: three types of annotation noise in LIDC-IDRI and the coarse noise in Cityscapes.
For LIDC-IDRI, we set $N_\mathrm{c}=1$ and used the B-mix gate, and for Cityscapes, we set $N_\mathrm{c}=2$ and disabled the B-mix gate.

To evaluate the quality of the corrected GT $\bm{Y}_\mathrm{c}$ and the accuracy of the trained DNN $\bm{f}$, the DSC with the clean GT was calculated and averaged over all images in the dataset.
The DNN was trained and tested five independent times to calculate the mean and standard error of the DSC.

An ablation study was also conducted, where we omitted the pre- or mid-correction to directly use $\bm{Y}_\mathrm{n}$ for the pre-train or to skip the mid-train, respectively.
To check the segmentation performance of SAM itself, we also tested a ``post-correction'', where the prediction by a trained DNN $\bm{f}$ was further corrected via SAM in the same way as the pre-correction.
The performance of the post-correction was evaluated by the DSC between the corrected prediction with the clean GT.
We also investigated combinations of these correction processes.

\section{Results}
\begin{figure}[t]
    \centerline{\includegraphics[width=\linewidth]{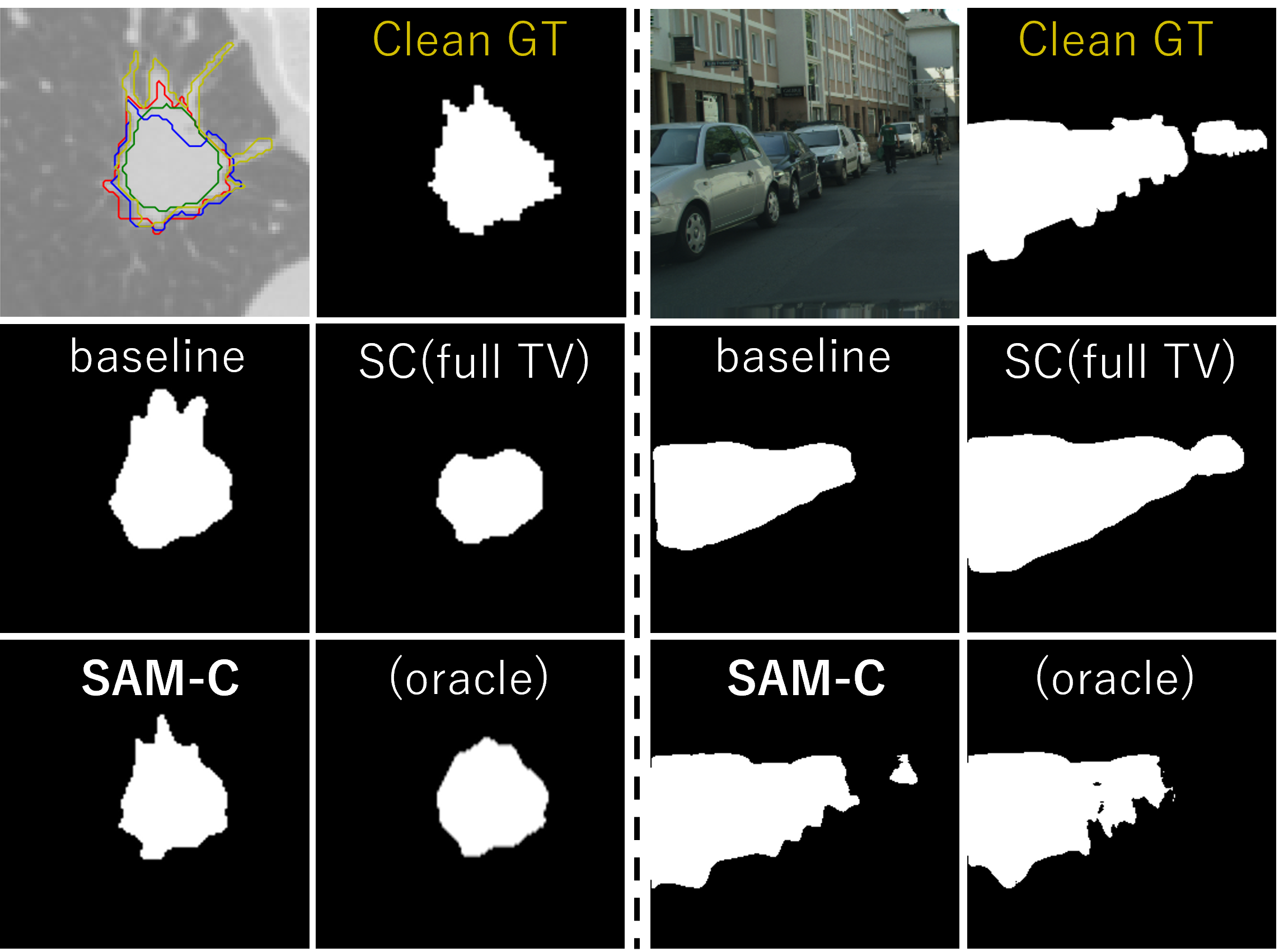}}
    \caption{Comparison of predictions by DNN.
    (Left) Results for LIDC-IDRI.
    All methods except the oracle used GTs with ``or noise''.
    Colored contours represent individual annotations.
    (Right) Results for Cityscapes.}
    \label{fig:2}
\end{figure}
Our proposed SAM-C demonstrated the highest DSC (see \tabref{tab:1}(a)).
While the oracle achieved the highest accuracy, SAM-C performed closest to this upper bound.
\figref{fig:2} shows that the DNN trained via SAM-C successfully segmented the region where annotation noise is particularly intense for LIDC-IDRI.
Fine structures such as the car wheel in Cityscapes were also well recognized through SAM-C.
Remarkably, SAM-C improved the performance even without additional well-curated GTs, which were required for SC to estimate the bias parameter.
Although SC with single TV could improve the accuracy of $\bm{f}$ for LIDC-IDRI, this was not always observed for Cityscapes, possibly due to overfitting the bias parameter.
This suggests that a too-small amount of well-curated GTs is insufficient for SC to perform robustly under a variety of conditions in practice.
Our SAM-C is, therefore, also advantageous in terms of the label acquisition cost.

The ablation study further underpinned the validity of our method.
We first found that pre-, mid-, and post-correction are all individually effective on their own (see \tabref{tab:1}(b)).
Intuitively, those correction processes can remove noise in annotations or predictions that deviate from the image texture.
This process does not require SAM to have medical domain knowledge. 
The mid-correction, in addition, leverages $\bm{f}(\bm{X})$ as prompts, where the trained DNN $\bm{f}$ acquires expert knowledge and mitigates semantically invalid noise.
The quality of $\bm{f}(\bm{X})$ as the prompt was then improved over $\bm{Y}_\mathrm{n}$, resulting in the superior performance of the mid-correction to the pre-correction.
Regarding the post-correction, the DSC between the corrected prediction and the clean GT strongly depends on the compatibility of SAM with the dataset.
The post-correction failed to achieve a higher DSC than the mid-correction, possibly because SAM itself did not have enough expertise to precisely segment objects.
We also confirmed that combining the pre- and mid-correction resulted in robustly outperforming SC in all experimental setups.
Although the combination of the mid- and post-correction achieved the highest DSC in some cases, the performance was sometimes rather inferior to SC, indicating that the post-correction is unstable across various experimental conditions.

In addition to the prediction accuracy of the trained DNN, we also evaluated the quality of the corrected GT mask $\bm{Y}_\mathrm{c}$ as a teacher label.
Accurate noise reduction from annotation labels is crucial not only to enhance the model accuracy but also to recover the noiseless learning process itself.
Specifically, we compared the quality of $\bm{Y}_\mathrm{n}$ and $\bm{Y}_\mathrm{c}$ based on the DSC with the clean GT in the training datasets.
For some cases using SC, we observed that the DSC of $\bm{Y}_\mathrm{c}$ was significantly below that of $\bm{Y}_\mathrm{n}$, even though the model accuracy is superior for SC to the baseline (see \tabref{tab:2}).
In contrast, our proposed SAM-C achieved a relatively robust reduction of annotation noise (see \tabref{tab:2}).
This could be because SAM recognizes the image texture to adaptively correct the mask, unlike SC, which refers to the typical bias length to simply expand or shrink the mask.
The precise correction of the teacher label by SAM-C may have led to the fine boundary structure of the prediction that was not apparent with SC (see \figref{fig:2}).

\begin{table}[t]
    \caption{Comparison between $\bm{Y}_\mathrm{n}$ and $\bm{Y}_\mathrm{c}$ regarding DSC with clean GT in the training dataset.
    In the numerals, blue, red, and black represent increase, decrease, and unchanged, respectively.}
    \label{tab:2}
    \centerline{\includegraphics[width=\linewidth]{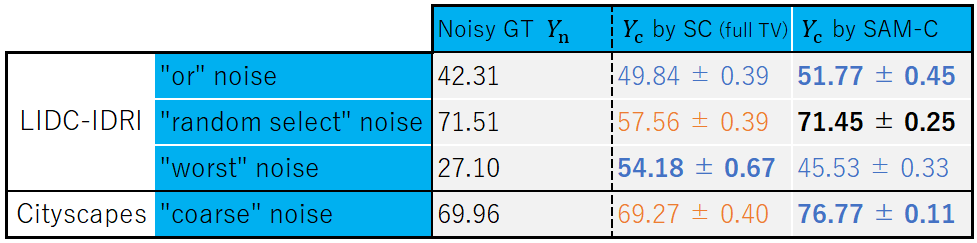}}
\end{table}

\section{Discussion and Conclusion}
We developed SAM-C to mitigate the annotation noise in teacher labels by utilizing SAM, a foundation model for image segmentation.
Even without additional well-curated GT masks, SAM-C outperformed the existing methods in terms of both the prediction accuracy of the trained DNN $\bm{f}$ and the precise reproducibility of the clean GT mask for training.
Our ablation study revealed that the mid-correction, which utilizes the prediction from the concurrently trained $\bm{f}$ to prompt SAM, played a critical role.

The potential limitations of SAM-C can be seen from the results in \tabref{tab:2}.
When using the ``worst'' noisy GT mask for LIDC-IDRI, the corrected GT $\bm{Y}_\mathrm{c}$ by SC was better than that of SAM-C in terms of the DSC with the clean GT.
This indicates that SAM-C might not perform well when the annotation deviates too much from the image texture, preventing SAM from precisely discriminating noisy labels.
However, the above limitation could be addressed by enhancing the capability of SAM---for example; we could simply use other foundation models such as HQ-SAM \cite{Ke2023} or SAM-Med2D \cite{Cheng2023}, which are being developed successively nowadays.
Likewise, the instability of the post-correction might be resolved by utilizing another foundation model compatible with the dataset domain.
Optimizing the prompt given to SAM, as implemented in AutoSAM \cite{Shaharabany2023}, might also be effective.

\section{Compliance with Ethical Standards}
Ethical approval was not required, as confirmed by the license attached with the open data we used.

\section{Acknowledgments}
The authors are employees of NEC corporation.

\bibliographystyle{IEEEbib}
\bibliography{paper}

\end{document}